\documentclass[10pt,twocolumn,letterpaper]{article}

\usepackage{cvpr}
\usepackage{times}
\usepackage{epsfig}
\usepackage{graphicx}
\usepackage{amsmath}
\usepackage{amssymb}

\usepackage[table,xcdraw]{xcolor}
\usepackage{dsfont}
\usepackage{array}
\usepackage{multirow}
\usepackage[caption=false]{subfig}
\usepackage{algorithm}
\usepackage{algorithmic}
\newcommand{\bd}[1]{\textbf{#1}}
\newcommand{\app}{\raise.17ex\hbox{$\scriptstyle\sim$}}

\newcolumntype{x}[1]{>{\centering\arraybackslash}p{#1pt}}

\newlength\savewidth\newcommand\shline{\noalign{\global\savewidth\arrayrulewidth
  \global\arrayrulewidth 1pt}\hline\noalign{\global\arrayrulewidth\savewidth}}
\newcommand{\tablestyle}[2]{\setlength{\tabcolsep}{#1}\renewcommand{\arraystretch}{#2}\centering\footnotesize}
\makeatletter\renewcommand\paragraph{\@startsection{paragraph}{4}{\z@}
  {.5em \@plus1ex \@minus.2ex}{-.5em}{\normalfont\normalsize\bfseries}}\makeatother

\usepackage[pagebackref=true,breaklinks=true,letterpaper=true,colorlinks,bookmarks=false]{hyperref}

\cvprfinalcopy % *** Uncomment this line for the final submission

\def\cvprPaperID{} % *** Enter the CVPR Paper ID here

\ifcvprfinal\fi

\begin{document}

%%%%%%%%% TITLE
\title{FastPose: Towards Real-time Pose Estimation and Tracking via Scale-normalized Multi-task Networks}

\author{Jiabin Zhang$^{1}$\thanks{The first two authors contributed equally to this work.}~~~~~Zheng Zhu$^{1}$\footnotemark[1]~~~~~Wei Zou$^{1}$~~~~~Peng Li$^{2}$~~~~Yanwei Li$^{1}$~~~~Hu Su$^{1}$~~~~Guan Huang$^{3}$\\
  $^1$Institute of Automation, Chinese Academy of Sciences, Beijing, China\\
  $^2$Beijing University of Posts and Telecommunications, Beijing, China\\
  $^3$Horizon Robotics, Beijing, China\\
  {\tt\scriptsize \{zhangjiabin2016,wei.zou,liyanwei2017, hu.su\}@ia.ac.cn, 
  zhengzhu@ieee.org,qqlipeng@bupt.edu.cn, 
  guanhuang@horizon.ai}
}

\maketitle
%\thispagestyle{empty}

%%%%%%%%% ABSTRACT
\begin{abstract}
Both accuracy and efficiency are significant for pose estimation and tracking in videos. State-of-the-art performance is dominated by two-stages top-down methods. Despite the leading results, these methods are impractical for real-world applications due to their separated architectures and complicated calculation. This paper addresses the task of articulated multi-person pose estimation and tracking towards real-time speed. An end-to-end multi-task network (MTN) is designed to perform human detection, pose estimation, and person re-identification (Re-ID) tasks simultaneously. To alleviate the performance bottleneck caused by scale variation problem, a paradigm which exploits scale-normalized image and feature pyramids (SIFP) is proposed to boost both performance and speed. Given the results of MTN, we adopt an occlusion-aware Re-ID feature strategy in the pose tracking module, where pose information is utilized to infer the occlusion state to make better use of Re-ID feature. In experiments, we demonstrate that the pose estimation and tracking performance improves steadily utilizing SIFP through different backbones. Using ResNet-18 and ResNet-50 as backbones, the overall pose tracking framework achieves competitive performance with 29.4 FPS and 12.2 FPS, respectively. Additionally, occlusion-aware Re-ID feature decreases the identification switches by 37\% in the pose tracking process.
\end{abstract}
%Then the three groups of results can provide useful information to the following pose tracking strategy. 
%%%%%%%%% BODY TEXT
\section{Introduction}
%-------------------------------------------------------------------------
% \begin{figure}[t]
% 	\begin{minipage}[t]{0.48\linewidth}
% 		\centering
% 		\includegraphics[width=1.02\linewidth]{pic/mAP.pdf}
% 		{\scriptsize \\(a) Inference speed and mAP}
% 	\end{minipage}
% 	\begin{minipage}[t]{0.48\linewidth}
% 		\centering
% 		\includegraphics[width=1.02\linewidth]{pic/MOTA.pdf}
% 		{\scriptsize (b) Inference speed and MOTA}
% 	\end{minipage}
%   \caption{(a):Inference speed and mAP performance on PoseTrack \cite{PoseTrackBenchmark} val set. Methods involved are PoseTrack \cite{Posetrack}, JointFlow \cite{JointFlow}, PoseFlow \cite{PoseFlow}, Det-and-track \cite{Det_and_Track}, FlowTrack \cite{MSRAPose}, and our FastPose framework with various backbones. (b): Inference speed and MOTA performance on PoseTrack val set.}
% \label{fig:speed}
% \end{figure}

\begin{figure}[t]
\centering
\includegraphics[width=1.0\linewidth]{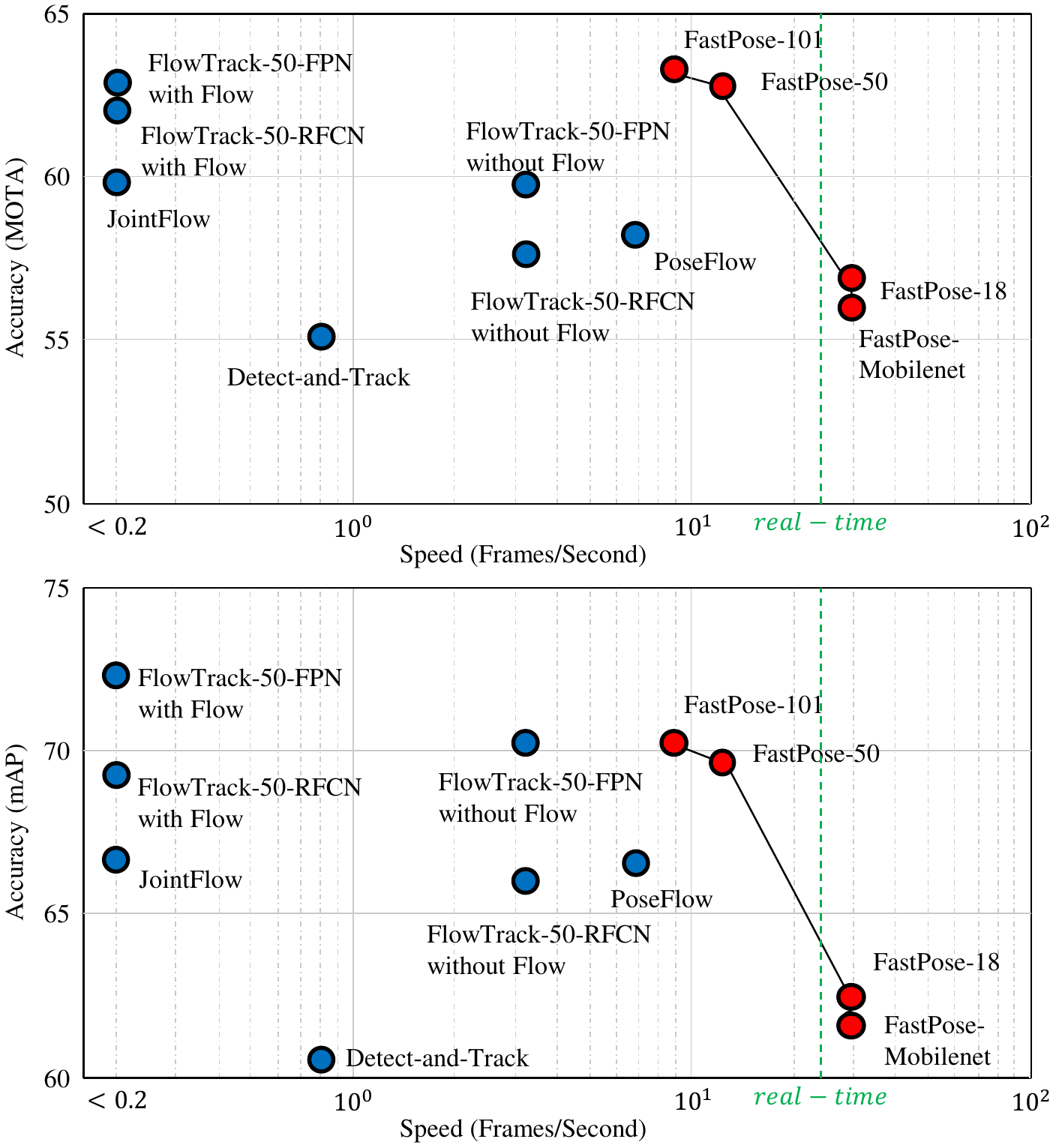}
   \caption{Top: Inference speed and MOTA performance on PoseTrack \cite{PoseTrackBenchmark} \texttt{val} set. Bottom: Inference speed and mAP performance on PoseTrack \texttt{val} set. Methods involved are PoseTrack \cite{Posetrack}, JointFlow \cite{JointFlow}, PoseFlow \cite{PoseFlow}, Detect-and-Track \cite{Det_and_Track}, FlowTrack \cite{MSRAPose}, and our FastPose framework with various backbones. A base 10 logarithmic scale is adopted for the \emph{x}-axis.}
\label{fig:speed}
\end{figure}
%-------------------------------------------------------------------------

Human pose estimation in images and articulated pose tracking in videos are of importance for visual understanding tasks \cite{zhou2014learning, mask}. Research community has witnessed a significant advance from single person \cite{PS,DPM,DeepPose,tompson2014joint,CPM,Hourglass,fppose} to multi-person pose estimation \cite{DeepCut,DeeperCut,OpenPose,GooglePose,CPN}, from static images pose estimation \cite{pishchulin2012articulated, mask} to articulated tracking in videos \cite{Posetrack,ArtTrack,JointFlow,Det_and_Track,PoseFlow,MSRAPose,zhang2019exploiting,li2018high,zhu2018distractor}. However, there are still challenging pose estimation problems in complex environments, such as occlusion, intense light and rare poses \cite{MPII,COCO,diagnosispose}. Furthermore, articulated tracking encounters new challenges in unconstrained videos such as camera motion, blur and view variants \cite{PoseTrackBenchmark}.
  
Previous pose estimation systems address single pre-located person, which exploit pictorial structures model\cite {PS,DPM} and following deep convolutional neural networks (DCNNs) approaches \cite{DeepPose,tompson2014joint,CPM,Hourglass,fppose}. Motivated by practical applications in video surveillance, human-computer interaction and action recognition, researchers now focus on the multi-person pose estimation in unconstrained environment. Multi-person pose estimation can be categorized into bottom-up \cite{DeepCut,DeeperCut,OpenPose} and top-down approaches \cite{GooglePose,CPN,mask,MSRAPose}, where the latter becomes dominant participant in recent benchmarks \cite{COCO,MPII}. Top-down approaches can be divided into two-stages methods and unified framework. Two-stages methods \cite{GooglePose,CPN,MSRAPose} firstly detect and crop persons from the image, then perform the single person pose estimation in the cropped person patches. Representative work of unified framework method is Mask R-CNN \cite{mask}, which extracts human bounding box and predicts keypoints from the corresponding feature maps simultaneously.

Generally, two-stages methods achieve the state-of-the-art results both on pose estimation and pose tracking tasks, beyond the performance of unified approach. We argue that the performance bottleneck of unified methods is caused by scale variation of the human. Specifically, the two-stages pose estimation methods are scale invariant. Based on the detection result of the first stage, the second stage only focuses on the task of keypoint detection on a fixed scale. Despite the leading performance, these methods can't perform real-time inference as their complex procedures, including human detection, cropping and scaling images, and pose estimation. In contrast, the unified frameworks can simply obtain the final multi-person pose estimation result from the original image in an end-to-end network. Unfortunately, the unified architecture destroys the scale invariance properties. Although many methods \cite{FPN,SFAD} have been proposed to alleviate scale variation problem in face detection or object detection area, there are few researches focusing on dealing with the scale variation in unified multi-person pose estimation. Recent researches \cite{SNIP,SNIPER} give an insight into the scale variation problem, but their inference speed suffers from multi-scale operation.

Different from multiple object tracking that focuses on instance identification assignment, pose tracking aims to address a more complex problem of articulated multi-person pose tracking in videos. Based on the bottom-up pose estimation methods, \cite{Posetrack, ArtTrack} construct spatial-temporal graphs between detected joints and solve a matching or energy optimization problem. However, high computing complexity of these methods makes it impractical for real-world applications. Based on the top-down pose estimation methods, \cite{MSRAPose} exploits flow-based pose similarity as metric and solves the matching problem in a greedy fashion. \cite{Det_and_Track} proposes a 3D extension of Mask R-CNN, which predicts the location of person tubes and corresponding poses simultaneously. In order to link these poses over time, they solve a bipartite graph matching problem based on intersection over union (IoU) metric. These simple tracking modules may encounter failure in challenging scenarios such as occlusions and crowds. Recent multiple object trackers \cite{wojke2017simple, bae2018confidence, yu2016poi, zhu2018online,feng2019multi,li2019state} prefer to use Re-ID features to maintain more robust track in these situation. However, Re-ID feature always becomes unreliable when the target is occluded. 
%几篇新的MOT

Based on the above analyses, this paper develops FastPose, a pose tracking framework which can perform pose estimation and tracking towards real-time speed. Specifically, we first build a multi-task network (MTN) which jointly optimizes three tasks simultaneously, including human detection, pose estimation, and person Re-ID. The three groups of outputs are utilized to perform pose tracking. Then a scale-normalized paradigm is proposed to alleviate the scale variation problem for the multi-task network. At last an occlusion-aware Re-ID strategy is designed for articulated multi-person pose tracking in video. To make better use of Re-ID features, we utilize the pose information to infer the occlusion state. 

The main contributions of this paper can be described as follows:

(1) Taking the person Re-ID features into account, we design an end-to-end multi-task network which performs human detection, pose estimation, and person Re-ID simultaneously. The network's outputs provide the necessary informations for the following pose tracking strategy.

(2) We propose a paradigm named scale-normalized image and feature pyramid (SIFP) for alleviating scale variation problem which is the performance bottleneck of unified top-down pose estimation methods. Based on image pyramid, we ignore extremely small and large objects to make sizes of objects uniformly distributed in the exact range. Combining feature pyramid networks (FPN) with the scale distribution can help the network to avoid multi-scale testing.

(3) Utilizing the outputs of our multi-task network, an occlusion-aware strategy is exploited to perform articulated multi-person pose tracking in videos. Specifically, the pose information is utilized to infer occlusion state and achieve the occlusion-aware Re-ID strategy which dramatically reduce the identification (ID) switches during tracking.

(4) In the experiments, our FastPose-18 (takes ResNet-18 as backbone) achieves real-time inference speed at 29.4 frames per image (FPS) while obtaining a mAP score of 63.1 and a MOTA score of 56.8. It is faster than other pose tracking approaches. Taking ResNet-50 as backbone, FastPose-50 achieves a fairly competitive performance at mAP of 69.7 and MOTA of 62.8 with a inference speed of 12.2 FPS. More detailed relationship between accuracy and inference speed of FastPose and other approaches is illustrated in Fig \ref{fig:speed}. Based on occlusion-aware Re-ID feature, our proposed tracking strategy achieves 37\% ID switches decrease over the tracking strategy without Re-ID feature.

\section{Related Works}

\subsection{Multi-person Pose Estimation in Image}
Pose estimation has underwent a long way as a basic research topic of computer vision. In recent years, motivated by practical applications, researchers switch focus from single person \cite{PS,DPM,DeepPose,tompson2014joint,CPM,Hourglass,fppose} to multi-person pose estimation. Different from single pre-located person, multi-person pose estimation can be categorized into bottom-up \cite{DeepCut,DeeperCut,OpenPose} and top-down approaches \cite{GooglePose,CPN,mask,MSRAPose}. CPN \cite{CPN} is the leading method on COCO 2017
keypoint challenge. It involves skip layer feature concatenation and an online hard keypoint mining step. \cite{MSRAPose} adopts FPN-DCN as the human detector and adds a few deconvolutional layers on single-person pose estimation network to improve the performance. These top-down methods achieve multi-person pose estimation by the two-stages process, including obtaining person bounding boxes by a person detector and predicting keypoint locations within these boxes. Besides, Mask R-CNN \cite{mask} builds an end-to-end framework and yields an impressive performance, but it is still behind these two-stages methods. We argue that the performance bottleneck of the unified approaches is caused by scale variation problem, which doesn't exist in above two-stages framework.

\begin{figure*}[ht]
\centering
\includegraphics[scale=0.40]{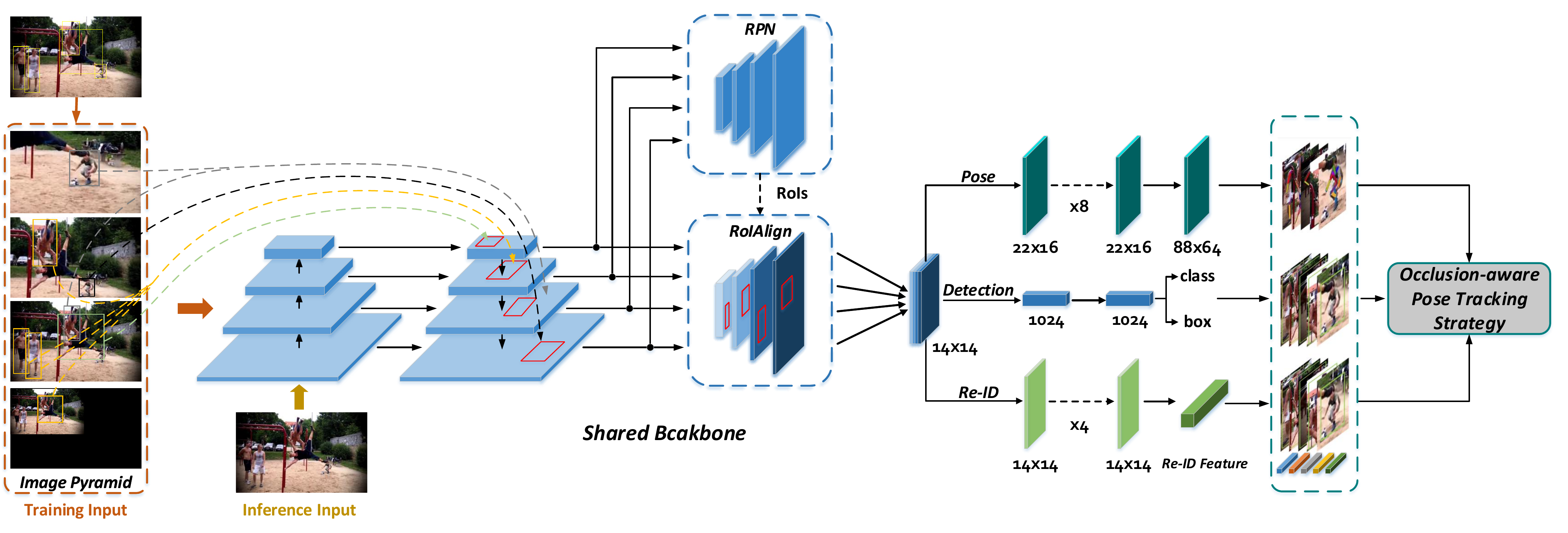}
\caption{The pipeline of the FastPose framework. In training process of multi-task network (MTN), a scale-normalized paradigm which exploits both image pyramid and feature pyramid is utilized to improve the distribution of objects' size. Utilizing the outputs of MTN, an occlusion-aware strategy performs pose tracking.}
\label{fig:net}
\vspace{-0.3cm}
\end{figure*}

\subsection{Multi-person Pose Tracking in Video}
Based on the multi-person pose estimation approaches described above, it is natural to extend them to multi-person pose tracking in video. Hence, the works of pose tracking can be also divided into bottom-up and top-down methods. In \cite{Posetrack, ArtTrack}, authors firstly estimate human pose with a bottom-up method, and then transform the problem into solving an energy minimizing function over a spatio-temporal graph constructed on the detected joints. \cite{JointFlow} proposes a model to predict Temporal Flow Fields (TTF) to formulate a similarity measure of detected joints. These similarities are used as binary potentials in a bipartite graph optimization problem in order to perform tracking of multiple poses. Based on the top-down pose estimation methods, \cite{Det_and_Track} proposes an extended Mask R-CNN and solves the bipartite graph matching problem based on IoU. \cite{MSRAPose} exploits flow-based pose similarity as metric and solves the matching problem in a greedy fashion. Based on the obtained pose of single person, \cite{PoseFlow} proposes to construct pose flow and perform pose flow non maximum suppression (NMS) to eliminate issues like ID switches. 

\subsection{Multi-task Learning}
Multi-task learning \cite{multitask,mtsurvey,benmulti,panoptic} has been used successfully in applications of natural language processing \cite{NLP,nlpmulti}, speech recognition \cite{speech}, computer vision \cite{Fast,facemulti,personmulti}. Especially in many computer vision tasks, the effectiveness of multi-task learning has been proved. Fast R-CNN \cite{Fast} and Faster R-CNN \cite{faster} jointly predict the class and the coordinates of objects in an image. Mask R-CNN \cite{mask} can efficiently detect objects in an image while simultaneously generating a high-quality segmentation mask for each instance. Similar with these methods, our approach shares the backbone network among all tasks, while keeping several task-specific output layers. This form has several advantages, for example, one end-to-end neural network saves much running time than several separated networks.

\subsection{Scale Invariant in Vision Tasks}
Large scale variation is one of major factors to influence the performance of many computer vision tasks like face detection, object detection and pose estimation. Many face detection approaches \cite{facescale1,facescale2,facescale3} have been proposed to learn representation that is invariant to scale. With the help of image pyramid \cite{pyramid}, some methods like DPM \cite{DPM} become more scale-robust. To address the problem that large strides of deep CNNs make small object detection very difficult, object detector \cite{deeplab, RFCN} use dilated/atrous convolutions to increase the resolution of the feature map. As the feature maps of higher layers have more semantic information but lower resolution, meanwhile these of the lower layers have high resolution. SDP \cite{SDP}, SSH \cite{SSH} and MS-CNN \cite{MS-CNN} make predictions of small objects on the lower layer and big objects on the higher layers respectively. Furthermore, methods like FPN \cite{FPN} and Mask-RCNN \cite{mask} propose a pyramidal representation and fuse adjacent scale feature maps to combine features which have semantic and detail informations. Besides, some methods, like SNIP \cite{SNIP} and SNIPER \cite{SNIPER}, propose advanced and efficient data argumentation methods to illustrate the scale variation problem. But they need multi-scale testing to achieve high performance, which harms the inference efficiency of network.

\section{Our Approach}
In this section, we discuss the proposed FastPose framework in details. The pipeline of the whole framework is illustrated in Fig. \ref{fig:net}. Given an original image as input, the multi-task network (MTN) can predict the bounding boxes, keypoint coordinates and Re-ID features in the scene. A scale-normalized image pyramid and feature pyramid (SIFP) paradigm is exploited to alleviate the scale variation problem of MTN. Following MTN, we propose an occlusion-aware pose tracking strategy for articulated multi-person pose tracking in video.

\subsection{The Multi-task Network (MTN)}
\label{sec:MTN}
The MTN adopts the similar unified procedure as Mask R-CNN. We first use a deep convolutional neural network (CNN) to transform original image to feature maps. A fully convolutional network, called a Region Proposal Network (RPN), is built upon these feature maps to propose candidate human bounding boxes. Based on the candidate boxes and their corresponding features extracted from the sharing feature maps, Mask R-CNN has two branches, one branch performs classification and bounding-box regression. Another branch outputs a binary mask for each human proposal, which can easily be extended to perform human pose estimation. To perform the task of extracting 128-d Re-ID features for each person in the image, we add a third branch that outputs the classification result of person's ID. 

\paragraph{Network Architecture:}Similar with Mask R-CNN, our proposed network can be instantiated with multiple architectures: (i) the \emph{backbone} network used for feature extraction over an entire image, and (ii) the \emph{head} networks for human detection (bounding-box classification and regression), pose estimation and person Re-ID that are applied separately to each RoI.

For the \emph{backbone} network, deeper architecture gains the effectiveness of extracted features, but brings longer training and inference time. To provide a trade-off between accuracy and speed when MTN is adopted in practical applications, we evaluate MobileNet-v2 \cite{mobile} and ResNet \cite{ResNet} with FPN \cite{FPN} of depth 18, 50 and 101 layers. 

For the pose estimation \emph{head} network, Mask R-CNN adopts a straightforward structure, which limits the precision of keypoints localization. MTN extends it to a more efficient structure. In Mask R-CNN,  $14\times14$ feature maps of 512 channels are extracted by RoIAlign for each proposal. In MTN, we utilize a padding operation to maintaining the ratio of the person in the $22\times16$ feature maps extracted by RoIAlign. After passing through a stack of $3\times3$ 512-d convolutional layers and two deconv layers, the spatial resolution is upsampled to $88\times64$.

For the person Re-ID \emph{head} network, a straightforward structure is adopted to classify each person's ID. Utilizing RoIAlign operation, a small feature map (\emph{e.g.} $512\times14\times14$) is extracted for each person proposal. Then it passes through a stack of $3\times3$ 256-d convolutional layers and transformed to a vector. A fully connected layer is used to summarize the vector into a 128-d feature vector. This 128-d Re-ID feature is utilized to compute the similarity metric between persons. In training process, taking the 128-d feature as input, another fully connected layer products a \emph{N}-d output, where \emph{N} depends on the ID scale of training dataset. For each person, the training target is a one-hot \emph{N}-d vector, and we minimize the \emph{cross-entropy} loss over a \emph{N}-way softmax output. To reduce computation complexity and bandwidth consumption, we only take top-128 person proposals into training process.  As this head network is based upon the backbone and RPN, so it need the training data composed by images within multi-person and corresponding ID annotation, like some person search datasets \cite{SSM, PRW}.

The MTN can provide necessary informations to the occlusion-aware strategy introduced in Sec. \ref{sec:strategy} to perform pose tracking. 

\subsection{Scale-normalized Image and Feature Pyramid (SIFP)}
As described above, MTN performs human detection, pose estimation and person Re-ID simultaneously. Different with two-stages methods which perform these tasks by separated networks, large scale variation across human instances is one of the main factors which influence the performance of our network, especially in pose estimation. Specifically, in the training and inference processes of two-stage methods, the scale of the input image for the single-person pose estimation network is fixed. However, MTN, a unified network, builds all the head networks upon the RoIs generated by RPN. This mechanism breaks scale invariant of MTN. So inspired by \cite{SNIP}, we develop a scale-normalized paradigm exploiting both image pyramid and feature pyramid (SIFP) to achieve enhanced scale invariance capability of MTN. 

In SIFP, we donate the scale \emph{s} of each object as $s=\sqrt{wh}$. Obviously, constraining \emph{s} of all the training objects to an intermediate scale range helps to reduce scale variation. By using an image pyramid where the original image is resized with a set of scaling factors ($\Omega={\{{\omega}_{i}\}}_{i=1}^{n}$), each object instance appears at several different scales and some of those appearances fall in the desired scale range. However, with $\omega > 1$ large objects become larger and with $\omega < 1$ small objects become smaller, which increases scale variation. Similar with \cite{SNIP}, SIFP only uses objects which fall in a certain scale range [$s_{l}$, $s_{u}$] as the training samples at each pyramid level. Additionally, images at a high resolution pyramid level are cropped to the same size of original image, without ignoring any valid objects. Images at a low resolution pyramid level are padded to the size of original image. In this way, all the object instances participate training, which preserves no-scale diversity and reduces scale variation in training the network. And fixed size at each pyramid level helps to utilize computing resources better.

If only using above extended image pyramid in training process, testing images also need to be resized to different scales with $\Omega$ to keep consistent. Because single scale testing would cause a large domain-shift due to the scale difference between training objects and testing objects. However, multi-scale testing would reduce the inference efficiency. To maxmize the inference speed without reducing performance, SIFP exploits FPN to tackle this dilemma. With FPN, anchors are defined to have areas of $S={\{{{s}_{a}}_{i}\}}_{i=1}^{5}$ on corresponding feature maps $P={\{{P}_{i}\}}_{i=1}^{5}$ respectively, for more details, please refer to \cite{FPN}. So the training objects in [$s_{l}$, $s_{u}$] are automatically distributed to different feature maps to assign labels to anchors. Each feature map only needs to focus on objects in a smaller scale range. In inference, test objects are distributed to different feature maps to be predicted. Due to FPN, MTN enhances scale invariance capability to alleviate the domain-shift brought by single scale testing. 

In conclusion, SIFP is a modified version of SNIP. Combining with FPN helps SNIP to avoid slower inference speed brought by multi-scale testing.  And our experiments in Sec. \ref{exp:ab} validate that even though testing on original image in order to meet towards real-time performance, our paradigm is very effective.

\subsection{Occlusion-aware Pose Tracking Strategy}
\label{sec:strategy}
% detect and track 等用什么策略 reid ，reid 问题
% 。。。， 最后说用了相同的策略
Based on the detection box, keypoints and Re-ID feature provided by MTN, pose tracking is performed by an occlusion-aware strategy. Strategies like \cite{Det_and_Track, highspeed} usually adopt IoU for linking tracks and ignore the appearance information, which fails to achieve competitive tracking result when the tracklets are occluded or in rapid movement. 
 Recently, Re-ID feature is widely adopted in multi-object tracking community as a stable appearance cue. However, Re-ID feature of the highly occluded target always contains invalid information, and may cause drift in tracking procedure. Therefore, inference of occlusion state is significant when adopting the Re-ID feature in complex scenarios. In this work, the occlusion-aware Re-ID feature is utilized as similarity metric to replace traditional IoU metric.
\subsubsection{Occlusion-aware Re-ID feature}
Human keypoints can be utilized to infer the occlusion state by the number of keypoints ($N_{valid}$) which are not occluded. And $N_{valid}$ is computed as:
\begin{equation}\label{equ:N_valid}
    N_{valid} = {\sum}_{i=1}^{N_k}\mathds{1}\{c_i > {\gamma}_{valid}\}
\end{equation}
where ${\gamma}_{valid}$ is the threshold for the confidence of keypoint $k_i$ to judge if $k_i$  is visible, $\mathds{1}$ equals 1 if the condition is true otherwise 0.

Re-ID feature is regarded valid when $N_{valid}$ is greater than the number threshold (${\theta}_{valid}$), which means that most of keypoints are visible and the target is not occluded, otherwise Re-ID feature is regarded invalid.
% 先说4再说3 ，用文本表示3， 详细解释3， 当大于一定数量认为feature无效

\subsubsection{Appearance feature of tracklet}
% 轨迹是由多帧构成的，需要维护以个轨迹外观特征
Tracklet consists of historical matched detections. Appearance feature of tracklet $f_{track}$ should be maintained carefully to make tracking procedure stable.
In some scenarios, target may move fast so their scale and pose change rapidly. The appearance feature will be updated if the Re-ID feature of matched detection is valid. % 被遮挡时不会更新
\subsubsection{Proposed similarity metric}
 Given Re-ID feature $f_d$ of detection $d$ and appearance feature $f_{track}$ of tracklet $track$, we adopt a integrated similarity metric $S$ containing position information and appearance information as:
 % 取消t-1，t以当前帧和维护的轨迹代替
 % however + IOU有什么问题
\begin{equation}\label{proposed cost metric}
    S = \theta_{pos} * IoU + (1 - \theta_{pos}) * \frac{min(dist(f_d, f_{track}), \sigma_{max})}{\sigma_{max}}
\end{equation}
where $\theta_{pos}$ controls the weight of IoU in $S$, and $dist(f_d, f_{track})$ means the Euclidean distance between feature $f_d$ and feature $f_{track}$. $\frac{min(dist(f_d, f_{track}), \sigma_{max})}{\sigma_{max}}$ is used to normalize $dist(f_d, f_{track})$ where ${\sigma}_{max}$ is the upper limit of Euclidean distance.

\subsection{Implementation Details}
There are some differences in network structure in details when we adopt various backbone networks for comprehensive experiments. As described in Sec. \ref{sec:MTN}, using a deeper backbone network (ResNet-50 or ResNet-101), the numbers of convolutional layers in pose estimation head and Re-ID head are 8 and 4 respectively. When using a smaller backbone (ResNet-18 or MobileNet-v2), they are changed to 4 and 2. 

\paragraph{Training:}The MTN needs three types of annotations corresponding to three head networks, including bounding box annotation, keypoint annotation, and human ID annotation. So the training process of pose tracking task is conducted on five datasets. COCO \cite{COCO} dataset provides bounding box and keypoint annotation. MPII \cite{MPII} and PoseTrack \cite{PoseTrackBenchmark} datasets are utilized for training pose estimation task. Person search datasets including SSM \cite{SSM} and PRW \cite{PRW} datasets are for training person Re-ID task. Image-centric training is adopted, for each image, the loss of unrelated tasks will not be propagated back.

The [$s_{l}$, $s_{u}$] is set as [16, 560] when SIFP is implemented. Only the objects whose $\sqrt{wh}$ fall in [16, 560] can be used to training in the image pyramid where the scaling factors are 2.0, 1.5, 1.0 and 0.75.%From our point of view, this suitable scale range makes feature maps of each level of FPN 

\paragraph{Inference:}At test time, for each frame images in videos, the proposal number provided by RPN is 1000 as in \cite{mask}. Human detection branch runs on these proposals. Utilizing non-maximum suppression, the highest scoring 100 detection boxes are fed into pose estimation and person Re-ID branches to obtain the heat maps of $K$ keypoints and 128-d Re-ID feature for each human boxes. After the inference of MTN, all the human boxes with their corresponding pose and Re-ID features are fed into our occlusion-aware tracking framework for articulated multi-person pose tracking in videos. We adopt a pose tracking strategy similar to \cite{Det_and_Track}. In the pose tracking strategy, ${\gamma}_{valid}$ is set as 0.2, $N_{valid}$ is set as 10, and $\theta_{pos}$ is set as 0.5.

%##################################################################################################
\begin{table*}[ht]\vspace{-3mm}
% subfloat a - Mask Architecture
\centering
\footnotesize
\subfloat[\textbf{Backbone Architecture and SIFP for Mask R-CNN}: Pose estimation results of Mask R-CNN without/with SIFP on different backbones. Among all the reported metrics, AP$^{kp}$ is the main metric of pose estimation on COCO dataset. We also report the inference speed of Mask R-CNN with different backbones on COCO dataset.\label{tab:ablation:mask backbone}]{
\tablestyle{2.5pt}{0.95}\begin{tabular}{r|x{30}x{30}x{30}x{30}x{30}x{30}x{36}x{36}x{30}}
  \emph{backbone} & AP$^\text{bb}_\text{\emph{person}}$ & AP$^\text{kp}$ & AP$^\text{kp}_{50}$ & AP$^\text{kp}_{75}$
 & AP$^\text{kp}_M$ &  AP$^\text{kp}_L$& Param(MB) & FLOPs(GB) &  \emph{speed}\\
\shline
  MobileNet-V2-FPN & 41.7 & 55.6 & 79.1 & 59.5 & 47.6 & 66.9 &  
 \multirow{2}{*}{22.73} & \multirow{2}{*}{33.6} &  \multirow{2}{*}{32.5}\\
  +SIFP & 43.9 & 57.9 & 81.1 & 62.1 & 51.4 & 67.8 & \\
\hline
  ResNet-18-FPN & 43.1 & 57.7 & 80.4 & 62.0 & 49.1 & 70.1  & 
 \multirow{2}{*}{32.38} & \multirow{2}{*}{63.4} &  \multirow{2}{*}{32.7}\\
  +SIFP & 45.3 & 60.1 & 82.1 & 64.3 & 53.4 & 70.9 & \\
\hline
  ResNet-50-FPN &  49.3 & 65.1 & 85.0 & 71.1 & 58.2 & 75.1  & 
 \multirow{2}{*}{51.78} & \multirow{2}{*}{109.8} &  \multirow{2}{*}{13.1}\\
  +SIFP &   52.9 & 67.5 & 85.8 & 73.6 & 62.4 & 75.8  & \\
\hline
  ResNet-101-FPN & 50.8 & 66.0 & 85.6 & 72.0 & 59.5 & 75.2 & 
 \multirow{2}{*}{67.48} & \multirow{2}{*}{147.8} &  \multirow{2}{*}{9.1} \\
  +SIFP & 53.9 & 68.3 & 86.5 & 74.4 & 63.2 & 76.4 & \\
%\hline
\end{tabular}}\hspace{3mm}

% subfloat b - MT Architecture
\subfloat[\textbf{Backbone Architecture and SIFP for FastPose}: Pose tracking results of FastPose without/with SIFP on different backbones. Among all the reported metrics, mAP and MOTA are two main metrics on PoseTrack dataset. We also report the inference speed of FastPose with different backbones on PoseTrack dataset.\label{tab:ablation:backbone}]{
\tablestyle{2.5pt}{0.95}\begin{tabular}[t]{r|x{21}x{21}x{36}x{36}x{26}}
  \emph{backbone}  & mAP & MOTA & Param(MB) & FLOPs(GB) & \emph{speed}\\
\shline
  MobileNet-V2-FPN & 60.9 & 52.1 & \multirow{2}{*}{32.73} & \multirow{2}{*}{38.2} & \multirow{2}{*}{28.6} \\
  +SIFP & 62.1 & 55.6 &  &  &  \\
 \hline
  ResNet-18-FPN & 62.0 & 53.9 &  \multirow{2}{*}{42.38} & \multirow{2}{*}{68.0} & \multirow{2}{*}{29.4} \\
  +SIFP & 63.1 & 56.8 &  &  & \\
 \hline
  ResNet-50-FPN & 69.0 & 60.1 & \multirow{2}{*}{62.98} & \multirow{2}{*}{116.8} & \multirow{2}{*}{12.2} \\
  +SIFP & 69.7 & 62.8 &  & & \\
 \hline
  ResNet-101-FPN & 69.5 & 60.6 & \multirow{2}{*}{78.68} &  \multirow{2}{*}{154.8} & \multirow{2}{*}{8.7} \\
  +SIFP & 70.3 & 63.2  & &  &  
\end{tabular}}\hspace{3mm}
% subfloat c -Pose Track strategy
\linespread{1.0}
\subfloat[\textbf{Pose tracking strategy}: Pose tracking results of three pose tracking strategy. They all base on the MTN-ResNet-50. The first strategy only utilizes IoU as similarity metric between persons. The second strategy uses Re-ID feature. The third one is our proposed strategy which uses occlusion-aware Re-ID feature. FP and FN are the numbers of false positive and false negative detected person. All the rows have the same FP and FN because their input comes from one MTN. IDS denotes ID switches.
\label{tab:ablation:strategy}]{
\tablestyle{2.5pt}{0.95}\begin{tabular}[t]{c|x{26}x{26}x{26}x{26}x{26}}
  Strategy & mAP & MOTA & FP & FN & \bd{IDS}\\
\shline
  \emph{IoU-only} & 69.7 & 62.2 & 1278.1 & 3704.0 & 243.1 \\
  \emph{Re-ID features} & 69.7 & 62.5 & 1278.1 & 3704.0 & 201.5 \\
\hline
   & - & \bd{+0.3} & - & - & \bd{-41.6} \\
 \hline
  \emph{occlusion-aware} & 69.7 & 62.8 & 1278.1 & 3704.0 & 153.9\\
\hline
  & - & \bd{+0.6} & - & - & \bd{-89.2}\\
% \multicolumn{4}{c}{~}\\
\end{tabular}}\hspace{3mm}

% subfloat d - Scale Normalization (ResNet-50-FPN)
\linespread{1.0}
\centering
\subfloat[\textbf{SIFP without/with FPN}:Results of utilizing SIFP with the backbone ResNet-50/ResNet-50-FPN. Baseline is ResNet-50-FPN without using SIFP. The second method is utilizing SNIP training strategy but testing on single scale, which means SIFP without FPN. The third one is the full SIFP. Metrics of pose estimation and pose tracking are reported simultaneously. \label{tab:ablation:sn-fpn}]{
\tablestyle{3.0pt}{0.95}\begin{tabular}{x{170}|x{26}x{26}x{26}x{26}x{26}x{26}|x{26}x{26}}
   & \multicolumn{6}{c|}{Mask R-CNN} & \multicolumn{2}{c}{FastPose} \\
\shline
  Method & AP$^\text{bb}_\text{\emph{person}}$ & AP$^\text{kp}$ & AP$^\text{kp}_{50}$ & AP$^\text{kp}_{75}$
 & AP$^\text{kp}_M$ &  AP$^\text{kp}_L$ & mAP & MOTA\\
\hline
  ResNet-50-FPN & 49.3 & 65.1 & 85.0 & 71.1 & 58.2 & 75.1 & 69.0 & 60.1  \\
  ResNet-50 + SNIP \cite{SNIP} training & 51.1 & 66.2 & 85.5 & 72.1 & 60.8 & 75.4 & 69.4 & 61.1 \\
  ResNet-50-FPN + SIFP& 52.9 & 67.5 & 85.8 & 73.6 & 62.4 & 75.8 & 69.7 & 62.8 
\end{tabular}}\hspace{3mm}
% subfloat e - Scale Normalization (ResNet-50-MS)
\centering
\subfloat[\textbf{SIFP v.s. MST}: Results of comparison SIFP with multi-scale training/testing. Baseline is ResNet-50-FPN without using SIFP. Metrics of pose estimation and pose tracking are reported simultaneously.\label{tab:ablation:sn-ms}]{
\tablestyle{3.0pt}{0.95}\begin{tabular}{x{170}|x{26}x{26}x{26}x{26}x{26}x{26}|x{26}x{26}}
   & \multicolumn{6}{c|}{Mask R-CNN} & \multicolumn{2}{c}{FastPose} \\
\shline
  Method & AP$^\text{bb}_\text{\emph{person}}$ & AP$^\text{kp}$ & AP$^\text{kp}_{50}$ & AP$^\text{kp}_{75}$
 & AP$^\text{kp}_M$ &  AP$^\text{kp}_L$ & mAP & MOTA\\
\hline
  ResNet-50-FPN & 49.3 & 65.1 & 85.0 & 71.1 & 58.2 & 75.1 & 69.0 & 60.1 \\
  ResNet-50-FPN + MS training\& testing  & 50.2 & 65.8 & 85.4 & 72.0 & 59.2 & 75.4 & 69.3 & 60.7\\
  ResNet-50-FPN + SIFP& 52.9 & 67.5 & 85.8 & 73.6 & 62.4 & 75.8 & 69.7 & 62.8
\end{tabular}}\hspace{3mm}

% main caption
\caption{\textbf{Ablations}. Pose estimation is achieved by Mask R-CNN, pose estimation and tacking in videos is achieved by MTN (FastPose). Mask R-CNN is trained on COCO \texttt{train}, tested on COCO\texttt{minival}. MTN is trained on COCO \texttt{train}, MPII \texttt{train}, PoseTrack \texttt{train}, SSM \texttt{train} and PRW \texttt{train}, tested on PoseTrack \texttt{val}.}
\label{tab:ablations}\vspace{-3mm}
\end{table*}
%##################################################################################################

\section{Experiments}
In this section, we perform thorough ablation experiments for both pose estimation and pose tracking tasks, and compare our FastPose framework with the state-of-the-art methods on PoseTrack \cite{PoseTrackBenchmark} dataset. In all the experiments, pose tracking task is evaluated on PoseTrack \cite{PoseTrackBenchmark} dataset. Pose estimation task is evaluated on 5k validation images (\texttt{minival}) of COCO \cite{COCO} dataset and PoseTrack \cite{PoseTrackBenchmark} dataset.

\subsection{Ablation Experiments}
\label{exp:ab}
Extensive of ablations are performed to analyze our approach, including different backbone architectures, different pose tracking strategies, and SIFP paradigm.

\paragraph{Backbone Architecture and SIFP for Mask R-CNN:}Table \hyperref[tab:ablation:mask backbone]{1(a)} shows our scale-normalized paradigm SIFP using in pose estimation with various backbones. AP$^{kp}$ is the main metric of pose estimation on COCO dataset. A deeper backbone has better performance. AP$^{kp}$ increase is 7.4 from ResNet-18 to Resnet-50. From ResNet-50 to Resnet-101, we obtain a small 0.8 improvement while FLOPs increases almost 30\%. So we adopt ResNet-50-FPN as the backbone for ablation studies in Table 1(c)-(d). Utilizing SIFP, AP$^{kp}$ is improved from 55.6 to 57.9, from 57.7 to 60.1, from 65.1 to 67.5, from 66.0 to 68.3 with the backbone of MobileNet-V2, ResNet-18, 50 and 101 respectively. One can find that all the architectures improve the pose estimation performance by using SIFP. 

\paragraph{Backbone Architecture and SIFP for FastPose:} As shown in Table \hyperref[tab:ablation:backbone]{1(b)}, our proposed FastPose also shows steady improvement by using deeper backbone models. mAP and MOTA are two main metrics on PoseTrack dataset. Using MobileNet-v2 or ResNet18 as the backbone, FastPose can achieve real-time pose tracking. Note that our inference speed doesn't grow with number of detected people, making it much more scalable to various scenes. Although FastPose-MobileNet-v2 has lower metric (62.1 on mAP and 55.6 on MOTA) than FastPose-18 (63.1 and 56.8), its properties make it particularly suitable for mobile applications. By using SIFP, mAP increases are 1.2, 1.1, 0.7 and 0.8, MOTA increases are 3.5, 2.9, 2.7 and 2.6 on listed backbones respectively. It proves SIFP can stably improve the performance of pose estimation and tracking on PoseTrack dataset. Pose estimation performance of FastPose on COCO dataset is reported in the supplementary material due to the page limit.

\paragraph{Occlusion-aware Re-ID feature:} Table \hyperref[tab:ablation:strategy]{1(c)} shows the effectiveness of Re-ID feature. Replacing IoU by Re-ID feature can make ID switches reduce 41.6 (from 243.1 to 201.5). Our proposed occlusion-aware strategy make a more remarkable improvement that reduces ID switches from 243.1 to 153.9 (37\%). Besides, we evaluate MTN on the person Re-ID dataset SSM and the mAP on SSM \texttt{test} is 89.38, which suggests that it is feasible to extract Re-ID features in MTN. Actually, this branch has a straightforward structure. More complex designs have the potential to improve performance but are not the focus of this work.

\paragraph{SIFP without/with FPN:} Table \hyperref[tab:ablation:sn-fpn]{1(d)} illustrates the results of combining SIFP with FPN and utilizing SIFP without FPN. The first row is the baseline that adopting ResNet-50-FPN as backbone without using SIFP. At the second row, utilizing SIFP without FPN is actually using SNIP's training strategy. This method introduces improvements of 0.9 AP$^{kp}$ for pose estimation, 0.4 mAP and 1.0 MOTA for pose tracking. At the third row, SIFP can obtain the improvement of 2.4, 0.7 and 2.8, all beyond the second row. So, our SIFP exploits FPN with SNIP's training strategy to improve pose estimation and tracking performance in single-scale testing.

\begin{table*}[th]
\tablestyle{4pt}{0.95}
\footnotesize
\begin{tabular}{l|lx{36}|x{22}x{22}x{30}x{20}x{20}x{20}x{20}x{20}x{20}x{20}}
\hline
\multirow{2}{*}{Method}& \multirow{2}{*}{Type} & \multirow{2}{*}{Detector} & \multirow{2}{*}{test set}&  mAP & mAP & mAP & mAP & mAP & mAP & mAP & \bd{mAP} & \bd{speed} \\
 &  &  &  &
 Head & Shoulder & Elbow & Wrist & Hip & Knee & Ankle & \bd{Total} & \bd{FPS}\\ 
\hline
 JointFlow \cite{JointFlow} & Bottom-up & - &  \texttt{val} & 
  & - & - & - & - & - & - & 66.7 & 0.2 \\ \cline{1-3}
 PoseFlow \cite{PoseFlow} & Top-down (2-stage) & SSD-512 &  \texttt{val} &
  66.7 & 73.3 & 68.3 & 61.1 & 67.5 & 67.0 & 61.3 & 66.5 &  6.7    \\
 FlowTrack-50-w/o Flow & Top-down (2-stage) & FPN-DCN  &  \texttt{val} &
  - & - & - & - & - & - & - & 69.3 &  3.0   \\
 FlowTrack-50 \cite{MSRAPose} & Top-down (2-stage) & FPN-DCN  &  \texttt{val} &
  79.1 & 80.5 & 75.5 & 66.0 & 70.8 & 70.0 & 61.7 & 72.4 &  0.2   \\
 FlowTrack-152-w/o Flow & Top-down (2-stage) & FPN-DCN  &  \texttt{val} &
  - & - & - & - & - & - & - & 72.9 &  3.2    \\
 FlowTrack-152 \cite{MSRAPose} & Top-down (2-stage) & FPN-DCN  &  \texttt{val} &
  81.7 & 83.4 & 80.0 & 72.4 & 75.3 & 74.8 & 67.1 & 76.7 &  0.2    \\
 Detect-and-Track \cite{Det_and_Track} & Top-down (end-end) & - &  \texttt{val} &
  67.5 & 70.2 & 62 & 51.7 & 60.7 & 58.7 & 49.8 & 60.6 & 0.8   \\ \cline{1-3}
 Ours:FastPose-18 & Top-down (end-end) & - &  \texttt{val} &
   76.7 & 73.6 & 62.2 & 51.1 & 63.6 & 58.0 & 48.7  & 63.1  & 29.4   \\ 
 Ours:FastPose-50 & Top-down (end-end) & - &  \texttt{val} &
  80.0 & 80.1 & 69.0 & 59.1 & 70.8 & 65.4 & 58.0 & 69.7 & 12.2   \\
 Ours:FastPose-101 & Top-down (end-end) & - &  \texttt{val} &
  80.0 & 80.3 & 69.5 & 59.1 & 71.4 & 67.5 & 59.4 & 70.3 & 8.7   \\
\hline
 JointFlow \cite{JointFlow} & Bottom-up & - &  \texttt{test} & 
  - & - & - & - & - & - & - & 63.3  & 0.2   \\
 PoseTrack \cite{PoseTrackBenchmark} & Bottom-up & - &  \texttt{test} &
  - & - & - & - & - & - & - & 59.4 &  -  \\ \cline{1-3}
 PoseFlow \cite{PoseFlow}& Top-down (2-stage) & SSD-512 &  \texttt{test} &
  64.9 & 67.5 & 65.0 & 59.0 & 62.5 & 62.8 & 57.9 & 63.0 & 6.7    \\
 FlowTrack-50 \cite{MSRAPose} & Top-down (2-stage) & FPN-DCN  &  \texttt{test} &
  76.4 & 77.2 & 72.2 & 65.1 & 68.5 & 66.9 & 60.3 & 70.0 & 0.2   \\
 FlowTrack-152 \cite{MSRAPose} & Top-down (2-stage) & FPN-DCN  &  \texttt{test} &
  79.5 & 79.7 & 76.4 & 70.7 & 71.6 & 71.3 & 64.9 & 73.9 & 0.2  \\
 Detect-and-Track \cite{Det_and_Track} & Top-down (end-end) & - &  \texttt{test} &
  - & - & - & - & - & - & - & 59.6 & 0.8  \\\cline{1-3}
 Ours:FastPose-18 & Top-down (end-end) & - &  \texttt{test} &
  74.1 & 71.2 & 60.1 & 45.3 & 59.8 & 54.2 & 46.5  & 60.2  & 29.4   \\ 
 Ours:FastPose-50 & Top-down (end-end) & - &  \texttt{test} &
  77.4 & 79.0 & 68.7 & 57.7 & 68.8 & 63.8 & 56.3 & 68.0 &  12.2  \\
 Ours:FastPose-101 & Top-down (end-end) & - &  \texttt{test} &
  77.8 & 79.4 & 69.5 & 58.2 & 69.7 & 65.6 & 57.6 & 68.9 &  8.7  \\
 \hline
\end{tabular}\vspace{1mm}
\caption{Multi-person Pose Estimation Performance on PoseTrack dataset.}\vspace{-3mm}
\label{tab:mAP}
\end{table*}

\begin{table*}[t]
\tablestyle{4pt}{0.95}
\begin{tabular}{l|lx{36}|x{22}x{22}x{30}x{22}x{22}x{22}}
\hline
\multirow{2}{*}{Method}& \multirow{2}{*}{Type} & \multirow{2}{*}{Detector} & \multirow{2}{*}{test set}&  \bd{MOTA} & MOTP & Prec & Rec & \bd{speed} \\
 &  &  &  &
 \bd{Total} & Total & Total & Total & \bd{FPS} \\ 
\hline
 JointFlow \cite{JointFlow} & Bottom-up & - &  \texttt{val} & 
 59.8 & - & 87.8 & 71.1  & 0.2 \\ \cline{1-3}
 PoseFlow \cite{PoseFlow} & Top-down (2-stage) & SSD-512 &  \texttt{val} &
 58.3 & 67.8 & 70.3 & 87.0 &  6.7    \\
 FlowTrack-50-w/o Flow & Top-down (2-stage) & FPN-DCN  &  \texttt{val} &
  59.8 & - & - & - & 3.2    \\
 FlowTrack-50 \cite{MSRAPose} & Top-down (2-stage) & FPN-DCN  &  \texttt{val} &
  62.9 & 84.5 & 86.3 & 76.0 & 0.2    \\
 FlowTrack-152-w/o Flow & Top-down (2-stage) & FPN-DCN  &  \texttt{val} &
  62.0 & - & - & - & 3.0    \\
 FlowTrack-152 \cite{MSRAPose} & Top-down (2-stage) & FPN-DCN  &  \texttt{val} &
  65.4 & 85.4 & 85.5 & 80.3 & 0.2   \\
 Detect-and-Track \cite{Det_and_Track} & Top-down (end-end) & - &  \texttt{val} &
  55.2 & 61.5 & 66.4 & 88.1 & 0.8    \\ \cline{1-3}
 Ours:FastPose-18 & Top-down (end-end) & - &  \texttt{val} &
   56.8 & 84.8  &  76.8  & 73.7 & 29.4   \\
 Ours:FastPose-50 & Top-down (end-end) & - &  \texttt{val} &
  62.8 & 85.2 & 88.8  & 73.1 & 12.2   \\
 Ours:FastPose-101 & Top-down (end-end) & - &  \texttt{val} &
  63.2 & 85.2 & 88.8 & 73.6 & 8.7   \\
\hline
 JointFlow \cite{JointFlow} & Bottom-up & - &  \texttt{test} & 
 53.1 & 82.6 &  69.7 & -  & 0.2   \\
 PoseTrack \cite{PoseTrackBenchmark} & Bottom-up & - &  \texttt{test} &
  48.4 & - & - & - &  -  \\ \cline{1-3}
 PoseFlow \cite{PoseFlow}& Top-down (2-stage) & SSD-512 &  \texttt{test} &
  51.0 & 16.9 & 71.2 & 78.9 & 6.7    \\
 FlowTrack-50 \cite{MSRAPose} & Top-down (2-stage) & FPN-DCN  &  \texttt{test} &
  56.4 & 45.5 & 81.0 & 75.7 & 0.2   \\
 FlowTrack-152 \cite{MSRAPose} & Top-down (2-stage) & FPN-DCN  &  \texttt{test} &
  57.6 & 62.6 & 79.4 & 79.9 & 0.2  \\
 Detect-and-Track \cite{Det_and_Track} & Top-down (end-end) & - &  \texttt{test} &
  51.8 & - & - & - & 0.8  \\\cline{1-3}
 Ours:FastPose-18 & Top-down (end-end) & - &  \texttt{test} &
  50.1 & 83.9  & 69.2  & 78.3 & 29.4   \\
 Ours:FastPose-50 & Top-down (end-end) & - &  \texttt{test} &
  56.6 & 84.7 & 78.1 & 78.2 &  12.2  \\
 Ours:FastPose-101 & Top-down (end-end) & - &  \texttt{test} &
  57.4 & 84.7 & 80.2 & 78.1 &  8.7  \\
 \hline
\end{tabular}\vspace{1mm}
\caption{Multi-person Pose Tracking Performance on PoseTrack dataset.}\vspace{-3mm}
\label{tab:MOTA}
\end{table*}

\paragraph{SIFP v.s. MST:} Multi-scale training and testing (MST) is another way to tackle scale variation problem. In Table \hyperref[tab:ablation:sn-ms]{1(e)}, we compare SIFP with MST. In multi-scale training process, the size of each training image is randomly scaled to one of 7 scales ((608, 1333), (640, 1333), (672, 1333), (704, 1333), (736, 1333), (768, 1333), (800, 1333)). The multi-scale testing result is a combination of testing results in all the same 7 scales. MST makes inference speed be 7 times slower, which is the price for improved metrics. However, it is worth noting that SIFP has the same inference time with the baseline, and increases AP$^\text{bb}_\text{\emph{person}}$, AP$^\text{kp}$, mAP and MOTA by 3.6, 2.4, 0.7 and 2.7 which are all higher than MST.

\subsection{Comparison with State-of-the-art}
we compare our FastPose framework with the state-of-the-art methods on PoseTrack Dataset \cite{PoseTrackBenchmark}, including PoseTrack \cite{PoseTrackBenchmark}, JointFlow \cite{JointFlow}, PoseFlow \cite{PoseFlow}, Detect-and-Track \cite{Det_and_Track} and FlowTrack \cite{MSRAPose}.

Table. \ref{tab:mAP} reports the results of pose estimation on PoseTrack dataset. Our FastPose-101 obtains mAP of 70.3 on \texttt{val} which outperforms the most methods, except FlowTrack. However, FlowTrack is a two-stages top-down method and has a significantly slower inference speed. Similarly using ResNet-101 as backbone, Detect-and-Track is almost 10 points behind FastPose-101 on mAP while its inference time is a tenth of ours. Other methods, no matter top-down or bottom-up approach, are all have lower mAP and slower speed than FastPose-50 or FastPose-101.

Table. \ref{tab:MOTA} reports the results of pose tracking on PoseTrack dataset. Our FastPose is also able to achieve competitive MOTA. On PoseTrack \texttt{val}, Only FlowTrack-152 with Flow has 65.4 MOTA higher than 63.2 of our FastPose-101. But its slower detector FPN-DCN and the optical flow estimation take much inference time, which causes the speed of FlowTrack-152 is only 0.2 FPS. Although using Flow and adopting FPN-DCN as human detector, FlowTrack-50 achieves MOTA of 62.9 which is still caught up by our FastPose-50 with MOTA of 62.8. On PoseTrack \texttt{test}, FastPose-50 and FastPose-101 achieve MOTA of 56.6 and 57.4, which are close to the state-of-the-art performance.

\paragraph{Timing}The last column of Table. \ref{tab:mAP} and \ref{tab:MOTA} shows the inference speed of compared methods. The speed is measured when FastPose is implemented by MXNet \cite{mxnet} on Intel Xeon E5-2620 @2.4GHz and NVIDIA TITAN X GPU. The inference time of FastPose comes from two aspects: MTN and tracking strategy. The inference speed of tracking strategy is 66.7 FPS. The speed of MTN is mainly depended on the two metrics of its architecture, including parameters (Param) and FLOPs calculated with setting the resolution of testing image as 600$\times$1000. Totally, the full inference speeds of FasePose are 28.6, 29.4, 12.2 and 8.7 FPS utilizing the four different backbones. Besides, the speed of FlowTrack is measured with the same hardware configuration to FastPose. The speed of PoseFlow (excludes pose inference time) is reported as 10 FPS in \cite{PoseFlow}, so we compute its final speed with the speed of RMPE referring to \cite{rmpe}. The speeds of other methods are cited from their papers. 

\section{Conclusion}
In this paper, we present FastPose, a fast and unified pose estimation and tracking framework, which utilizes a multi-task network (MTN) to integrates three tasks together. An occlusion-aware strategy following MTN performs pose tracking. Besides, a paradigm named Scale-normalized Image and Feature Pyramid (SIFP) is designed to deal with severe scale variation widely existed in unified pose approaches. In ablation studies, we prove the stable improvements brought by MTN, SIFP and occlusion-aware strategy. Moreover, with different configurations, FastPose can achieve real-time inference or competitive performance, which is helpful to adopt pose tracking in actual scenarios.

{\small
\bibliographystyle{ieee}
\bibliography{egbib}
}

\end{document}